\documentclass{article}

    \PassOptionsToPackage{numbers, compress}{natbib}

    \usepackage[preprint]{neurips_2025}

\usepackage[utf8]{inputenc} 
\usepackage[T1]{fontenc}    
\usepackage{url}            
\usepackage{booktabs}       
\usepackage{amsfonts}       
\usepackage{nicefrac}       
\usepackage{microtype}      

\usepackage{amsmath}
\usepackage{graphicx}
\usepackage{array}
\usepackage{multirow}
\usepackage[table]{xcolor}
\usepackage{tcolorbox}
\usepackage[hidelinks]{hyperref}

\title{RARL: Improving Medical VLM Reasoning and Generalization with Reinforcement Learning and LoRA under Data and Hardware Constraints}

\author{%
Tan-Hanh Pham$^{1,\dag}$, Chris Ngo$^2$\\
\hphantom{text}\\ 
$^1$Harvard Medical School, USA, $^2$Knovel Engineering Lab, Singapore\\
\hphantom{text}\\
$^\dag$\small{Corresponding author}
}

\begin{document}

\maketitle

\begin{abstract}
The growing integration of vision-language models (VLMs) in medical applications offers promising support for diagnostic reasoning. However, current medical VLMs often face limitations in generalization, transparency, and computational efficiency-barriers that hinder deployment in real-world, resource-constrained settings. To address these challenges, we propose a Reasoning-Aware Reinforcement Learning framework, \textbf{RARL}, that enhances the reasoning capabilities of medical VLMs while remaining efficient and adaptable to low-resource environments. Our approach fine-tunes a lightweight base model, Qwen2-VL-2B-Instruct, using Low-Rank Adaptation and custom reward functions that jointly consider diagnostic accuracy and reasoning quality. Training is performed on a single NVIDIA A100-PCIE-40GB GPU, demonstrating the feasibility of deploying such models in constrained environments. We evaluate the model using an LLM-as-judge framework that scores both correctness and explanation quality. Experimental results show that RARL significantly improves VLM performance in medical image analysis and clinical reasoning, outperforming supervised fine-tuning on reasoning-focused tasks by approximately 7.78\%, while requiring fewer computational resources. Additionally, we demonstrate the generalization capabilities of our approach on unseen datasets, achieving around 27\% improved performance compared to supervised fine-tuning and about 4\% over traditional RL fine-tuning. Our experiments also illustrate that diversity prompting during training and reasoning prompting during inference are crucial for enhancing VLM performance. Our findings highlight the potential of reasoning-guided learning and reasoning prompting to steer medical VLMs toward more transparent, accurate, and resource-efficient clinical decision-making. \href{https://github.com/Hanhpt23/MedicalImagingReasoning}{\textcolor{teal}{Code}} and \href{https://huggingface.co/datasets/Hanhpt23/RARL}{\textcolor{teal}{data}} are publicly available.
\end{abstract}

\section{Introduction}
The rapid progress in large language models (LLMs) has significantly reshaped artificial intelligence (AI), enabling machines to perform complex language understanding and generation tasks with remarkable fluency \cite{brown2020language, touvron2023llama, touvron2023llama2, dubey2024llama,qwen3}. This advancement has laid the groundwork for the development of multimodal models, particularly vision-language models, which combine visual perception with linguistic reasoning. These innovations have led to growing interest in domains where visual and textual data naturally intersect-most notably in medicine. In clinical settings, interpreting diagnostic images alongside textual patient data is central to effective decision-making. As such, medical VLMs have emerged as promising tools to support radiologists and clinicians in tasks such as automated abnormality detection, report generation, and image-based QA \cite{huang2023visual,moor2023med,tu2024towards,zhang2023biomedclip,zhang2023biomedgpt}. For instance, models like Llava-Med \cite{li2023llava} and Med-PaLM M \cite{tu2024towards} have demonstrated proficiency in analyzing chest X-rays to identify pathologies or generating structured reports from ultrasound images. These advancements hold significant promise for augmenting clinical workflows, particularly in high-stakes settings where rapid and accurate diagnoses are critical. 

Despite these advances, the development of medical VLMs is constrained by several practical challenges. First, high-performing models, such as LLaVA-Med \cite{li2023llava} or Med-PaLM M \cite{tu2024towards}, rely on large-scale datasets (e.g., MIMIC-CXR \cite{johnson2019mimic} or PMC-15M \cite{zhang2023pmc}) and extensive computational resources, often requiring clusters of high-end GPUs for training. Such requirements are prohibitive for smaller healthcare institutions, low-resource clinical settings, or global health initiatives \cite{hosny2018artificial}. Second, many medical VLMs are optimized for specific tasks or datasets, resulting in poor generalization to unseen or diverse clinical scenarios. For instance, a model trained on chest X-rays from a single hospital may struggle to analyze images from different imaging protocols or patient demographics \cite{pan2025medvlm,lai2025med}. However, despite the potential of these systems, several challenges remain. Notably, many existing models struggle with generalization across diverse medical datasets, demand computationally intensive training regimes, and most critically, lack transparent reasoning capabilities essential for clinical trust and accountability \cite{amann2020explainability}.

To address this problem, recent efforts have explored reasoning-focused techniques, such as chain-of-thought (CoT) prompting, which encourages models to articulate intermediate steps before delivering answers \cite{wei2022chain}. For instance, CoT has been applied in VLMs to improve performance on medical VQA tasks by generating structured diagnostic explanations, such as identifying radiological signs before concluding a diagnosis \cite{liu2024medcot}. For dataset, the medical reasoning datasets are also developed to enhance the reasoning of the medical VLMs such as \cite{pham2025silvar}. In terms of the training techniques, reinforcement learning (RL) have demonstrated its potential over self-supervised learning to enhance reasoning in medical VLMs following the success of Deepseek \cite{yao2024deepseek}.

Recently, RL-based approaches have gained attention for optimizing both diagnostic accuracy and reasoning depth \cite{chen2024reinforcement}. Studies like Med-R1 \cite{lai2025med} and MedVLM-R1 \cite{pan2025medvlm} have improved performance in tasks such as detecting abnormalities in chest radiographs by refining model outputs through reward-driven optimization. A key innovation is Group Relative Policy Optimization (GRPO), an advanced RL framework that balances multiple objectives, including diagnostic correctness, explanation clarity, and response conciseness \cite{liu2024grpo}. 

Despite these efforts, medical VLMs remain limited to multiple-choice or closed-ended questions; beyond that, the models struggle to generalize to other types of questions or to data that are out of distribution. In addition, while MedVLM-R1 \cite{pan2025medvlm} was trained using reinforcement learning (RL), it does not generalize well across different question types. Furthermore, although recent advances have improved the interpretability of medical VLMs, their application in resource-constrained settings remains underexplored. In terms of evaluation, current methods for assessing reasoning models primarily focus on final answer accuracy, often overshadowing the reasoning process and leaving models ``ill-equipped'' to provide the detailed explanations required for clinical validation \cite{amann2020explainability}.

To address these limitations, we propose a resource-efficient training pipeline that enhances both reasoning and diagnostic capabilities under constrained data and hardware conditions. Our method combines a novel Reasoning-Aware Reinforcement Learning (\textbf{RARL}) strategy with Low-Rank Adaptation (\textbf{LoRA}) fine-tuning. Using Qwen2-VL-2B, a lightweight VLM, we train our model on a single NVIDIA A100-PCIE-40GB GPU. For evaluation, we employ an LLM-as-judge framework that holistically assesses both answer correctness and the quality of reasoning.

Our contributions are summarized as follows:
\begin{itemize}
    \item We introduce \textbf{RARL}, a reinforcement learning framework to effectively improve reasoning quality, interpretability, and generalization in small VLMs under resource constraints, addressing a critical gap in medical AI deployment.
    \item Training on a single GPU demonstrates feasibility for real-world clinical settings.
    \item The 27\% gain on unseen datasets (e.g., VQA-RAD) highlights the framework’s adaptability.
\end{itemize}
\section{Methodology}

\subsection{Dataset}

To bridge the gap between general-purpose vision-language understanding and domain-specific medical reasoning, we hypothesize that explicit reasoning supervision is essential. To support this, we curated a small, quality reasoning dataset from Silvar-Med \cite{pham2025silvar}, which consists of 866 samples annotated by three physicians, covering MRI (22.4\%, 194 samples), CT scans (16.5\%, 143 samples), and chest X-rays (61.1\%, 529 samples). The dataset is split into training and testing sets: 716 samples for training (162 MRI, 122 CT, 432 X-ray) and 150 samples for testing (32 MRI, 21 CT, 97 X-ray). Unlike prior works that primarily focus on predicting short final answers or multiple choices \cite{chen2024huatuogpt}, our study is explicitly experimented for reasoning explainability. In the dataset, each question is paired with a detailed reasoning-based answer and an original answer, where the model must not only predict the correct label but also provide a clinically coherent explanation supporting the decision. The data samples are shown in Table \ref{tab:reasoning_dataset}. In addition to our curated dataset, we use VQA-RAD \cite{lau2018dataset}, SLAKE (English)\cite{liu2021slake}, VQA-Med 2019 \cite{ImageCLEFVQA-Med2019} to evaluate the generalization of our approaches.

\begin{table}[ht]
\centering
\caption{Side-by-side comparison abnormal detection.}
\begin{tabular}{>{\arraybackslash}m{6.5cm} >{\arraybackslash}m{6.5cm}}
\toprule
\rowcolor{gray!10}
\textbf{Sample 1: Abnormality detection} & \textbf{Sample 2: Abnormality detection} \\
\midrule

\textbf{Question}: \small{Does this image look abnormal?} \vspace{0.2cm} \newline
\textbf{Answer}: <think> Yes, the image appears abnormal due to the presence of irregularities in the lung fields and potential signs of pathology. The overall lung structure may show unusual densities or patterns that deviate from normal anatomical features. These findings could indicate conditions such as infection, inflammation, or other pulmonary issues that warrant further investigation </think> <answer> yes </answer>. 

\vspace{0.2cm} \textbf{Short answer}: Yes. 
\vspace{0.2cm} \newline
\includegraphics[width=5cm]{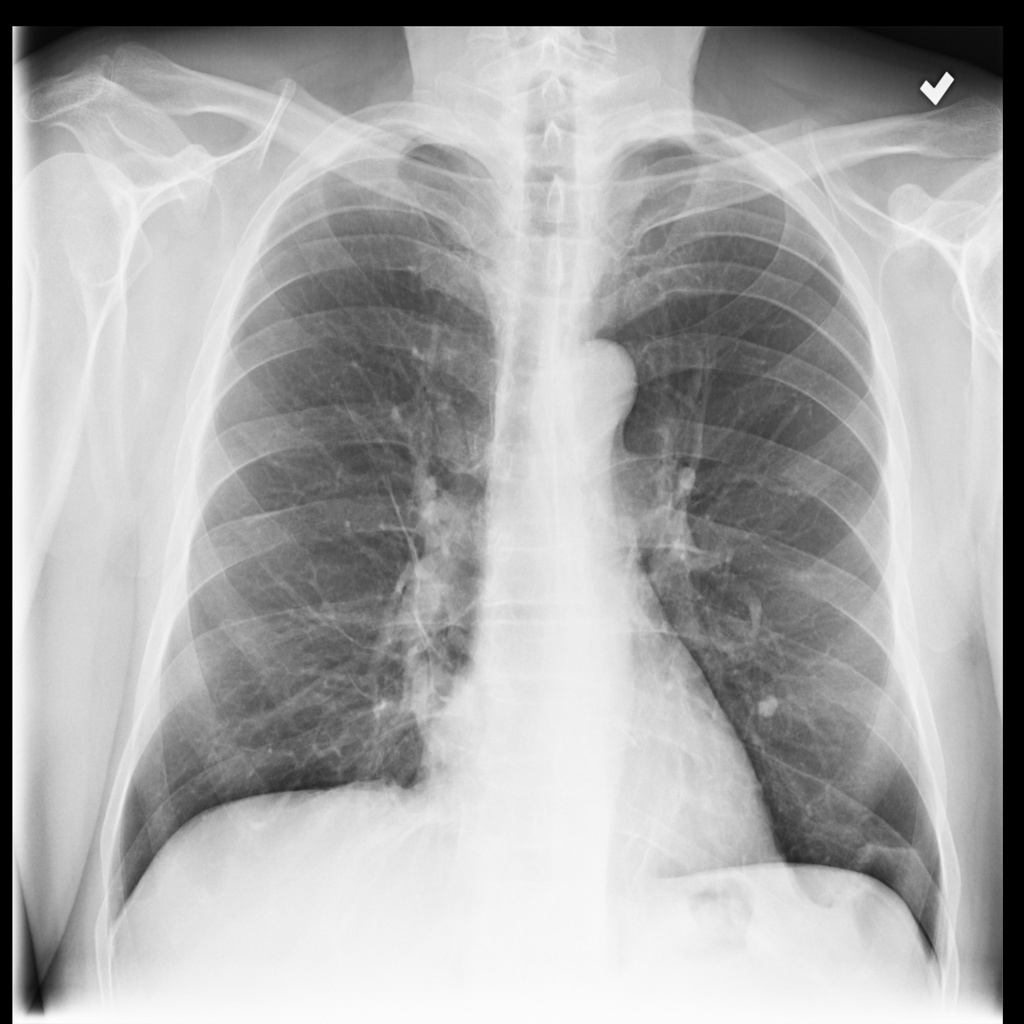} &

\textbf{Question}: \small{Is the lung healthy?} \vspace{0.2cm} \newline
\textbf{Explainable Answer}: <think> The lung appears to show signs of pathology, indicated by the presence of abnormal structures and potential lesions. These findings suggest that the lung is not functioning optimally and may be affected by a disease process. Further evaluation and clinical correlation are necessary to determine the specific condition affecting the lung. Overall, the imaging suggests that the lung is not healthy </think> <answer> no </answer>. 

\vspace{0.2cm} \textbf{Short answer}: No. 
\vspace{0.2cm} \newline
\includegraphics[width=5cm]{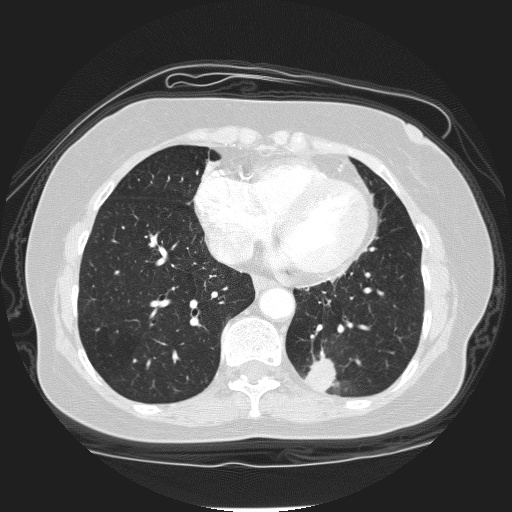} \\

\bottomrule
\end{tabular}
\label{tab:reasoning_dataset}
\end{table}

\subsection{Fine-Tuning Pipeline}
\label{finetuning_pipeline}
\subsubsection{Fine-Tuning with RL GRPO}
We fine-tune the model using GRPO, a reinforcement learning algorithm designed to enhance reasoning ability by sampling multiple outputs per input and optimizing based on group-normalized relative advantages \cite{shao2024deepseekmath}. GRPO eliminates the need for a separate value network by estimating baselines directly from sampled rewards, reducing computational overhead for resource-constrained settings.

Given an input question $q$, we sample a group of $G$ candidate outputs $\{v_1, v_2, \ldots, v_G\}$ from the old policy $\pi_{\theta_{\text{old}}}$. The policy $\pi_\theta$ is then optimized by maximizing the following objective:

\begin{equation}
\begin{aligned}
\mathcal{J}_{\text{GRPO}}(\theta) =\; & 
\mathbb{E}_{q \sim p(Q), \{v_i\}_{i=1}^G \sim \pi_{\theta_{\text{old}}}(\cdot \mid q)} \Bigg[\frac{1}{G} \sum_{i=1}^G \min \left(r_i(\theta) A_i,\,\operatorname{clip}(r_i(\theta), 1 - \epsilon, 1 + \epsilon) A_i \right) \\
& \qquad\qquad
- \beta \, \mathrm{D}_{\mathrm{KL}} \left( \pi_\theta(\cdot \mid q) \,\|\, \pi_{\text{ref}}(\cdot \mid q) \right)
\Bigg],
\end{aligned}
\end{equation}
where the advantage $A_i$ is computed from a group of rewards $\{r_1, r_2, \ldots, r_G\}$:
\begin{align}
A_i = \frac{r_i - \text{mean}(\{r_1, r_2, \ldots, r_G\})}{\text{std}(\{r_1, r_2, \ldots, r_G\})},
\end{align}
and the KL-divergence term is:
\begin{align}
\text{D}_{\text{KL}}(\pi_\theta \parallel \pi_{\text{ref}}) = \frac{\pi_{\text{ref}}(v_i \mid q)}{\pi_\theta(v_i \mid q)} - \log \frac{\pi_{\text{ref}}(v_i \mid q)}{\pi_\theta(v_i \mid q)} - 1.
\end{align}
Here, $\epsilon$ controls the clipping range for stability, and $\beta$ is the KL penalty coefficient to prevent excessive divergence from the reference policy $\pi_{\text{ref}}$ (pretrained model). The reward $r_i$ is computed using a composite rule-based function that evaluates the output format (e.g., correct use of <think> and <answer> tags), the correctness of the initial reasoning, and final prediction.

\subsubsection{Reward Functions}
\label{sec:RewardFunctions}
We design the reward function which is composed of components to encourage both clinical correctness and explainable reasoning.

\[
\texttt{<think>}~\text{\textcolor{purple}{Reasoning Steps}}~\texttt{</think>} \quad \texttt{<answer>}~\text{\textcolor{purple}{Final Answer}}~\texttt{</answer>}
\]

\textbf{Format Reward}:  
This term enforces the output structure by checking for the presence of <think>...</think> and <answer>...</answer> tags. A full reward (1.0) is given if the output matches the exact format. If only partial tagging is present, a small bonus reward (0.1) is assigned. Outputs without any tags receive zero reward. This component ensures that the model consistently produces interpretable and structured responses.

\textbf{Length Reward}:  
This term incentivizes sufficiently detailed explanations by rewarding longer outputs. The reward is proportional to the output length, measured by token count, scaled linearly with a maximum cap of 1.0. Specifically, a reward of $\texttt{min}(0.001 \times \text{token count}, 1.0)$ is assigned. This prevents extremely short or trivial completions that lack sufficient reasoning.

\textbf{Accuracy Reward}:  
This term measures the correctness of the model's final answer enclosed within the <answer>...</answer> tags. For multiple-choice or binary (yes/no) questions, a binary reward (1.0 for correct, 0.0 otherwise) is assigned. For open-ended questions, BERTScore F1 \cite{bert-score} is used to compute the semantic similarity between the predicted and reference answers, yielding a continuous reward between 0.0 and 1.0.

\textbf{Reasoning Reward}:  
In addition to the rewards described above, we introduce a \textbf{Reasoning Aware Reward} that specifically incentivizes the generation of meaningful, structured reasoning within the <think>...</think> tags.  
This reward is motivated by our observation that the base model, pretrained on general-domain data, has limited medical knowledge and thus requires additional guidance to develop coherent clinical reasoning pathways. This observation is similar to the real case when you recognize an MRI image of the brain but do not know how to logically explain why certain features suggest a tumor versus a benign cyst. Without explicit reasoning guidance, a model (or even a human learner) might memorize visual patterns ("bright spot = tumor") without understanding critical underlying features such as mass effect, surrounding edema, or contrast enhancement patterns. Thus, to cultivate clinically meaningful reasoning rather than shallow pattern recognition, it is essential to incentivize structured thought processes during adaptation to the medical domain.  


\subsection{Low-Rank Adaptation}

To further reduce the number of training parameter for hardware resources, we leverage the low-rank adaptation finetuning. LoRA is applied with rank $r=8$, $\alpha=16$, and target modules \texttt{["q\_proj", "k\_proj", "v\_proj", "o\_proj"]} to reduce trainable parameters, enabling single-GPU training. The vision encoder is frozen to focus optimization on language components, reducing computational overhead. We train the model for 5 epochs, with checkpoints saved every 100 steps to capture peak performance and mitigate instability. In addition, we use Flash Attention to accelerate faster training and inference \cite{dao2022flashattention}. 

\section{Experiments and Results}
\label{sec.experiment}
\subsection{RL and LoRA on General-Domain VQA under Data Constraints}
\label{sec.proof-of-method}
Before adapting our proposed training pipeline to the medical domain, we first validate its effectiveness on general-domain tasks using the Cambrian Vision-Centric Benchmark (CV-Bench) \cite{tong2024cambrian}. This step is motivated by the fact that the base model is extensively pretrained on general-domain data, making it both appropriate and fair to evaluate changes in its performance. CV-Bench evaluates diverse visual reasoning skills such as object counting, spatial relationship inference, and depth ordering, making it an ideal benchmark to isolate the effects of our training pipeline without the added complexity of domain adaptation. Our pipeline combines LoRA for parameter-efficient fine-tuning and GRPO-based reinforcement learning. It is important to note that we do not use the reasoning-aware reward in this experiment, as the dataset lacks reasoning annotations. To assess its robustness under data scarcity, we fine-tune the model using 500, 1,000, and 5,000 training samples, and compare three training setups: (1) standard supervised fine-tuning (SFT), (2) GRPO-based RL without LoRA \cite{zhou2025r1}, and (3) our proposed method: GRPO-based RL with LoRA. 

\begin{table}[h]
\centering
\small
\caption{Comparison of different training strategies on CV-Bench under varying sample sizes. "RL w/ LoRA" is evaluated at multiple data scales to demonstrate scalability.}
\label{tab:cvbench-results}
\begin{tabular}{lcccccc} 
\toprule
\textbf{Method} & \begin{tabular}{@{}c@{}} \textbf{Training} \\ \textbf{Samples}\end{tabular} & \begin{tabular}{@{}c@{}} \textbf{Count} \\ \textbf{Acc.} (\%)\end{tabular} & \begin{tabular}{@{}c@{}} \textbf{Relationship} \\ \textbf{Acc.} (\%)\end{tabular} & \begin{tabular}{@{}c@{}} \textbf{Depth} \\ \textbf{Acc.} (\%)\end{tabular} & \begin{tabular}{@{}c@{}} \textbf{Distance} \\ \textbf{Acc.} (\%)\end{tabular} & \begin{tabular}{@{}c@{}} \textbf{Total} \\ \textbf{Acc.} (\%)\end{tabular} \\
\midrule
\multicolumn{7}{l}{\textit{Experiment from literatures}}\\
\addlinespace
RL only (w/o LoRA) \cite{zhou2025r1} & 15,000  & 69.54\% &  61.84\%  & 66.50\%  & 65.50\% & 66.03\% \\

\midrule
\addlinespace
\multicolumn{7}{l}{\textit{Our experiments with Qwen2-VL-2B-Instruct}}\\
\addlinespace
SFT & 500 &  59.26\%   &  52.61\%  &  34.34\%  &  38.33\%  &  47.19\%  \\
SFT & 1,000 & 59.90\%   &  51.54\% & 46.67\%  & 46.00\%  &  51.67\% \\
SFT & 5,000 &  60.27\%  &  65.54\% &  45.67\% &  42.83\% & 54.28\%  \\
\cmidrule{2-7}
RL w/ LoRA  & 500 &  58.74\% & 54.19\%  & 38.00\%  &  41.23\% & 48.04\% \\
RL w/ LoRA & 1,000 & 60.30\%  & 57.23\%  &  48.33\% & 47.47\%  &  53.33\% \\
RL w/ LoRA & 5,000 &  63.89\% &  68.61\% &  52.84\% &  50.83\% &  59.04\%\\
\cmidrule{2-7}
RL only (w/o LoRA) & 5,000  & 64.43\% &  65.65\%  & 50.35\%  & 55.71\% & 59.29\% \\
\bottomrule
\end{tabular}
\end{table}

The results is shown in Table~\ref{tab:cvbench-results}, in which we use the same prompt in the training process for evaluation (reasoning prompt). The experiment demonstrates that, with smaller training datasets (500 and 1,000 samples), our approach (RL w/ LoRA) outperforms the SFT method on the CV-Bench, indicating better performance in limited data scenarios. Moreover, when trained with a larger dataset (5,000 samples), the performance of the model using RL and LoRA remains competitive with the model trained without LoRA. This suggests that our approach is not only effective for smaller datasets but also holds promise for real-world scenarios where data may be limited, showcasing its scalability and robustness.
 
\begin{tcolorbox}
[colback=black!3!white,colframe=purple!60!white,boxrule=0.3mm,title=, left=0mm,right=0mm]
\textbf{Observation 1}: The proposed method, RL with LoRA, outperforms standard SFT and performs competitively compared to RL without LoRA when trained on small datasets.
\end{tcolorbox}

\subsection{General-to-Medical Domain Adaptation}
\label{sec:general-to-medical}

After investigating the effectiveness of our training pipeline on the CV-Bench, we extend our training pipeline to the medical domain.  
Specifically, we fine-tune the model on our curated reasoning-focused medical training dataset and test its performance on our reasoning test set, along with several widely used medical VQA benchmarks: VQA-RAD \cite{lau2018dataset}, SLAKE (English) \cite{liu2021slake}, and VQA-Med 2019 \cite{ImageCLEFVQA-Med2019}.

Unlike our test set, the other benchmarks mainly consist of non-reasoning tasks in open-ended or closed-form formats. Due to potential mismatches between the model's free-form outputs and the ground-truth answers, direct string-matching is insufficient, so we adopt an \textbf{LLM-as-Judge} strategy, including GPT-4o mini and Gemini 1.5 Flash, to evaluate the reasoning and accuracy of the model’s generated answers relative to the ground truth. For the reasoning, we only evaluate the prediction of the model with the ground truth on our dataset since the other datasets do not content explanation.

\begin{table}[h]
\centering
\small
\caption{Evaluation of the model on different medical VQA datasets with different finetuning methods.}
\label{tab:llm-judge-results}
\begin{tabular}{lcccccc}
\toprule
\textbf{Dataset} & \multicolumn{2}{c}{\textbf{GPT-4o mini}} & \multicolumn{2}{c}{\textbf{Gemini 1.5 Flash}} & \multicolumn{2}{c}{\textbf{Human Evaluation}} \\ \cmidrule{2-3} \cmidrule{4-5} \cmidrule{6-7}
\textbf{} & \begin{tabular}{@{}c@{}} \textbf{Reasoning} \\ (\%)\end{tabular}
 & \begin{tabular}{@{}c@{}} \textbf{Final} \\ (\%)\end{tabular} & \begin{tabular}{@{}c@{}} \textbf{Reasoning} \\ (\%)\end{tabular} & \begin{tabular}{@{}c@{}} \textbf{Final} \\ (\%)\end{tabular} & \begin{tabular}{@{}c@{}} \textbf{Reasoning} \\ (\%)\end{tabular} & \begin{tabular}{@{}c@{}} \textbf{Final} \\ (\%)\end{tabular} \\
\midrule
\multicolumn{7}{l}{\textit{Our experiment using SFT}}\\
\addlinespace
Ours & 67.57\% & 64.86\% & 65.54\% & 62.21\% & 63.52\% & 60.81\% \\
VQA-RAD &  —  & 26.16\% &  — & 22.37\% & — & — \\
SLAKE  &  —  & 43.14\% &  —  & 42.89\% & — & — \\
VQA-Med 2019&  —  & 13.80\% &  —  & 11.43\% & — & — \\
Path-VQA  &  —  & 9.09\% &  —  & 8.88\% & — & — \\

\midrule

\multicolumn{7}{l}{\textit{Our experiments with the propsed pipeline: RL + LoRA}}\\
\addlinespace
Ours & 68.91\% & 61.49\% & 66.89\% & 60.81\% & 64.86\% & 62.84\% \\
VQA-RAD &  —  & 43.90\% &  —  & 42.13\% & — & — \\
SLAKE  &  —  & 51.64\% &  —  & 52.02\% & — & — \\
VQA-Med 2019&  —  & 49.80\% &  —  & 49.20\% & — & — \\
Path-VQA  &  —  & 24.26\% &  —  & 23.05\% & — & — \\
\bottomrule
\end{tabular}
\end{table}

Table~\ref{tab:llm-judge-results} illustrates the results of the model across multiple medical VQA datasets. We compare the result between our pipeline and SFT finetuning method. The experiments show that the model achieved up to 64.86\% reasoning accuracy and 62.84\% final answer accuracy based on human evaluation on our test set, demonstrating the model’s ability to follow logical inference processes. In addition, both GPT-4o mini and Gemini 1.5 Flash show consistent scoring trends, validating the use of LLM-as-Judge frameworks for semantic evaluation. Notably, we observe a gap between the reasoning and final answer scores. This discrepancy arises because the model often demonstrates valid reasoning - including correct intermediate steps - but may still produce an answer that does not exactly match the ground truth. As shown in Table~\ref{tab:reasoning_and_final_acc}, The model's prediction accurately identifies the potential conditions affecting the right lung, including pneumonia, which aligns with the ground truth description, but it also introduces other possibilities like lung cancer without sufficient justification, slightly reducing its relevance. This highlights the importance of evaluating not only the final answer but also the reasoning process behind it.

\begin{table}[ht]
\centering
\small
\caption{Model prediction and reasoning.}
\begin{tabular}{>{\arraybackslash}m{10cm} >{\centering\arraybackslash}m{3cm}}
\toprule
\rowcolor{gray!10} \multicolumn{2}{l}{\textbf{Comparison between final prediction and the model reasoning.}} \\ \midrule

\textbf{Question}: What disease is/are shown on the right of lung?\\ \vspace{-1.0cm}\hspace{2cm}

\textbf{Short answer}: \textcolor{orange}{Pneumonia} 

\textbf{Grounth truth}: The image shows \textcolor{orange}{pneumonia} in the right lung, characterized by areas of increased opacity due to fluid accumulation or inflammation. This condition can result from various infectious agents, leading to symptoms such as cough, fever, and difficulty breathing. The presence of such opacities indicates that the lung tissue is affected, which is typical in pneumonia cases. &  \includegraphics[width=3cm]{xmlab314.jpg} \\ \midrule
\multicolumn{2}{p{13.5cm}}{\textbf{Model reasoning}: The image is a chest X-ray, and the right lung appears to have a darker, more opaque area compared to the left lung. This is indicative of a lung opacity, which can be caused by various conditions such as \textcolor{orange}{pneumonia}, \textcolor{purple}{lung cancer, or other lung diseases.} It is important to consult with a healthcare professional for a proper diagnosis and treatment plan. }\\ 
\multicolumn{2}{p{12cm}}{\textbf{Model final prediction}: pneumonia, lung cancer and other lung diseases.}\\ 
\bottomrule
\end{tabular}
\label{tab:reasoning_and_final_acc}
\end{table}

\begin{table}[ht]
\centering
\small
\caption{Prediction of the model on an unseen dataset using our pipeline.}
\begin{tabular}{>{\centering\arraybackslash}m{4cm} >{\arraybackslash}m{9cm}}
\toprule
\rowcolor{gray!10}
\textbf{Image} & \textbf{Explanation} \\
\midrule
\includegraphics[width=3.5cm]{xmlab13.jpg} & 
\textbf{Question}: Does the \textcolor{orange}{liver} look normal?
\vspace{0.1cm}

\textbf{Ground truth}: Yes.
\vspace{0.1cm}

\textbf{Model prediction}: \textcolor{orange}{The liver appears to have a normal size and shape, with no visible lesions or abnormalities in the surrounding structures.} The internal texture seems consistent, indicating healthy liver tissue. Overall, there are no signs of cirrhosis, fatty liver, or other pathological changes. Therefore, the liver can be considered normal based on this imaging.\\
\bottomrule
\end{tabular}
\label{tab:liver_normal_case}
\end{table}

For generalization, Table~\ref{tab:llm-judge-results} illustrates that our pipeline consistently outperforms SFT across public benchmarks, showing 5–35\% gains in accuracy including VQA-RAD, SLAKE, and VQA-Med 2019. 
The lowest performance was observed on Path-VQA, an out-of-distributed, large-scale dataset with highly diverse visual and question content. Table~\ref{tab:liver_normal_case} provides a qualitative example, illustrating that the model generates accurate and well-explained predictions for a normal liver case on the SLAKE benchmark.

While our method performs well across medical VQA datasets, we observe two key limitations:
\begin{itemize}
\item \textbf{Hallucinated or weak reasoning:}
The model sometimes generates generic or poorly grounded explanations, indicating a need for better alignment between the question and reasoning.
\item \textbf{Short answer challenges:}
For binary or factual questions, the model may over- or under-generate content, struggling to balance brevity with sufficient context.
\end{itemize}

\begin{tcolorbox}
[colback=black!5!white,colframe=purple!50!white,boxrule=0.3mm,title=, left=0mm,right=0mm]
\textbf{Observation 2}: 
(1) The proposed method outperforms the SFT fine-tuning approach in both human and LLM-as-judge evaluations on the reasoning dataset. 
(2) The model demonstrates better generalization to unseen datasets when trained with our pipeline under limited data conditions. 
(3) We show that evaluating only the final answer is insufficient, as the model may perform meaningful reasoning but still fail to produce the optimal answer.
\end{tcolorbox}

\subsection{Thinking Initialization and Prompting Diversity}
\label{sec:thinking-init}

\begin{figure}[h]
\centering
\includegraphics[width=0.7\linewidth]{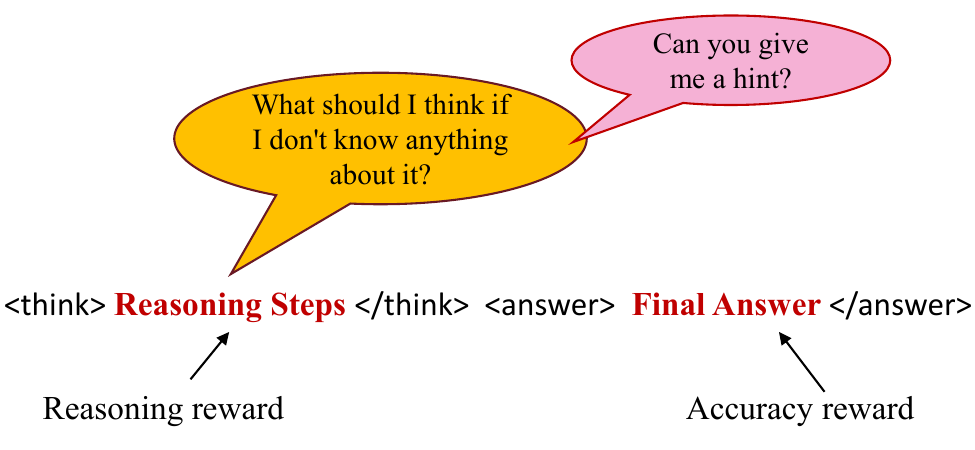}
\caption{Overview of \textbf{RARL}, which incentivizes both clinically faithful intermediate reasoning and accurate final answers.}
\label{fig:thinking_initialization}
\end{figure}

As shown in Sections~\ref{sec.proof-of-method} and~\ref{sec:general-to-medical}, our pipeline for small-scale medical VLMs shows strong performance under limited data and hardware constraints. However, we observe that such models still tend to produce hallucinated or shallow answers in complex medical scenarios, suggesting that optimizing solely for final answer accuracy is insufficient for clinical reliability. To address this challenge, we introduce two complementary strategies: \textbf{Reasoning-Aware Reinforcement Learning or RARL}, and \textbf{Prompting Diversity}, designed to enhance the model’s reasoning consistency, clinical robustness, and adaptability to a wide range of medical queries in resource-limited settings.

In the \textbf{RARL} framework, for training samples annotated with reasoning, the model is guided to generate structured intermediate thoughts enclosed within \texttt{<think>}...\texttt{</think>} tags before producing the final answer. A reasoning reward, proposed in section \ref{sec:RewardFunctions}, evaluates the clinical coherence and informativeness of these intermediate steps, while a separate reward targets final answer correctness. For samples without reasoning annotations, we apply a standard reward based only on the final answer. This dual-reward design encourages a the model get some certain knowledge by initializing thinking and free thinking style making the model getting better faster with some useful knowledge in the medical domain.

\textbf{For prompting diversity}, to enhance the model's generalization across varied clinical scenarios, we introduce diverse prompting styles during training. Rather than using a fixed question-answer format, we expose the model to a range of prompt types that reflect real-world clinical queries:
        
\begin{itemize}
    \item \textit{Explanation-Required Answers}: These prompts encourage the model to engage in thorough clinical reasoning before providing an answer, which should include relevant justifications or supporting information.
    \item \textit{Short-Form Answers}: These prompts encourage concise factual or yes/no answers, accompanied by minimal but contextually relevant justification.
    \item \textit{Open-Ended Answers}: These prompts promote free-form reasoning to handle unstructured or ambiguous clinical scenarios.
\end{itemize}

To further investigate the model’s performance during inference, we evaluate the impact of different prompting techniques: \textbf{reasoning prompting} and \textbf{non-reasoning prompting}, defined as follows:
        
\begin{itemize}
    \item \textit{Reasoning prompting}: "A conversation between a User and a Medical Assistant. The User asks a question about the image, and the Assistant solves it. The Assistant first thinks through the reasoning process step-by-step, then provides the answer to the User.”
    \item \textit{Non-reasoning prompting}: "A conversation between a User and a Medical Assistant. The Assistant provides the answer to the User."
\end{itemize}

\begin{table}[h]
\centering
\small
\caption{Performance evaluation of the model on medical VQA datasets using LLM-as-Judge scoring with RARL and Prompting Diversity.}
\label{tab:RARL_results}
\begin{tabular}{lcccccc}
\toprule
\textbf{Dataset} & \multicolumn{2}{c}{\textbf{GPT-4o mini}} & \multicolumn{2}{c}{\textbf{Gemini 1.5 Flash}} & \multicolumn{2}{c}{\textbf{Human Evaluation}} \\ \cmidrule{2-3} \cmidrule{4-5} \cmidrule{6-7}
\textbf{} & \begin{tabular}{@{}c@{}} \textbf{Reasoning} \\ (\%)\end{tabular}
 & \begin{tabular}{@{}c@{}} \textbf{Final} \\ (\%)\end{tabular} & \begin{tabular}{@{}c@{}} \textbf{Reasoning} \\ (\%)\end{tabular} & \begin{tabular}{@{}c@{}} \textbf{Final} \\ (\%)\end{tabular} & \begin{tabular}{@{}c@{}} \textbf{Reasoning} \\ (\%)\end{tabular} & \begin{tabular}{@{}c@{}} \textbf{Final} \\ (\%)\end{tabular} \\
\midrule
\multicolumn{7}{l}{\textit{Experiment reasoning prompting in inference}}\\
\addlinespace
Ours & 70.27\% & 64.86\% & 68.92\% & 64.86\% & 70.94\% & 65.54\% \\
VQA-RAD & — & 45.73\% & — & 44.00\% & — & — \\
SLAKE  & — & 56.38\% & — & 54.71\% & — & — \\
VQA-Med 2019& — & 51.18\% & — & 50.47\% & — & — \\
Path-VQA  & — & 25.33\% & — & 24.78\% & — & — \\
\midrule
\multicolumn{7}{l}{\textit{Experiment none-reasoning prompting in inference}}\\
\addlinespace
Ours & 67.56 & 61.49 & 65.54\% & 61.49\% & 65.64\% & 60.81\% \\
VQA-RAD & — & 40.54\% & — & 40.21\% & — & — \\
SLAKE  & — & 52.67\% & — & 52.55\% & — & — \\
VQA-Med 2019 & — & 47.99\% & — & 46.31\% & — & — \\
Path-VQA  & — & 24.00\% & — & 23.72\% & — & — \\
\bottomrule
\end{tabular}
\end{table}

Similar to the experiment in Section \ref{sec:general-to-medical}, we evaluate our approach on multiple medical VQA benchmarks. As summarized in Table~\ref{tab:RARL_results}, combining RARL with Prompting Diversity significantly boosts performance across datasets. On our reasoning test set, the model increase by around 4.5\% with LLM-as-judge and human evaluation compared the the case without RARL. Compared to the baseline, final answer scores assessed by GPT-4o mini improved by 4.17\% on VQA-RAD, 9.18\% on SLAKE, 2.77\% on VQA-Med 2019, and 4.41\% on Path-VQA. From the results, the integration of RARL and Prompting Diversity significantly improves the model’s reasoning behavior compared to conventional RL approaches. RARL’s dual-reward mechanism ensures that the model prioritizes clinically relevant reasoning, reducing instances of hallucination and generic responses observed in Section~\ref{sec:general-to-medical}. For example, on reasoning-focused tasks in our test set, the model demonstrates stronger alignment with ground-truth explanations, as evidenced by higher LLM-as-Judge scores and human evaluation. In addition, there is a gap between the reasoning and non-reasoning prompts. The performance of the model decrease by 4.01\%, 5.16\%, and 7.78\% evaluated by GPT 4o mini, Gemini1.5 Flash and human, respectively. Despite the performance improvement, challenges remain, particularly in generalizing to out-of-distribution data (Path-VQA benchmark), where the model produces less robust reasoning, indicating a need for broader training data or additional domain-specific fine-tuning.

\begin{tcolorbox}
[colback=black!5!white,colframe=purple!50!white,boxrule=0.3mm,title=, left=0mm,right=0mm]
\textbf{Observation 3}:  
(1) The experiment shows that reasoning-aware training (\textbf{RARL}) and prompting diversity not only improve performance on reasoning tasks but also help the model internalize more clinically relevant knowledge that generalizes effectively to different types of data.  
(2) Reasoning prompting during the inference stage is crucial for guiding the model to produce correct answers.
\end{tcolorbox}

\section{Conclusion}

In this work, we introduced a Reasoning-Aware Reinforcement Learning (RARL) framework along with several strategies to enhance the clinical reasoning capabilities of medical vision-language models (VLMs) under constrained data and hardware resources. By fine-tuning a lightweight model, Qwen2-VL-2B-Instruct, using Low-Rank Adaptation (LoRA) and custom reward functions that jointly optimize diagnostic accuracy and reasoning quality, our approach achieves strong performance while remaining computationally efficient.

We evaluated our method using an LLM-as-judge framework that considers both the correctness and quality of explanations. RARL significantly outperforms supervised fine-tuning by 7.78\% in reasoning-focused tasks and generalizes better to unseen datasets, with a 27\% improvement over SFT and a 4\% gain over traditional reinforcement learning approaches. These results underscore the value of incorporating reasoning supervision and reward design into medical VLMs, offering a practical path toward more interpretable and effective clinical decision support systems. Furthermore, our findings highlight the importance of jointly evaluating the reasoning process and the final answer in reasoning-intensive tasks. We also show that prompting diversity during inference can complement RARL in guiding small VLMs toward clinically aligned reasoning.

Despite these advancements, challenges remain. The model hallucinates or struggles with generalization to out-of-distribution data, indicating the need for more diverse training data and robust domain adaptation methods. Future work will focus on addressing these limitations by expanding training coverage and exploring advanced prompting techniques to further reduce hallucinations and improve generalization.

\section*{Limitation}

Despite the benefits of our proposed pipeline for improving medical VLM reasoning and generalization under limited data and hardware constraints, our work also has some limitations, including the dataset, LLM-as-judge bias, and medical clinician evaluations.


\bibliographystyle{unsrtnat}
\bibliography{ref}

\newpage
\appendix
\label{sec:appendix}
\section{Prompts for Training VLMs}

\begin{tcolorbox}
[colback=black!0!white,colframe=black!0!white,boxrule=0.3mm,title=, left=0mm,right=0mm]
\textbf{Explanation-Required Answer Prompt:}

\texttt{prompt = f"You are a Medical Assistant. Carefully analyze the medical image and the User’s question. Think through the problem step by step using clinical reasoning and relevant visual cues. Present your thinking process clearly within <think> </think> tags, and then provide a well-justified answer within <answer> </answer> tags. User: {question} Assistant: Let me solve this step by step."}

\textbf{Short-Form Answer Prompt:}

\texttt{prompt = f"You are a Medical Assistant. Read the User's question and the associated medical image. Provide your thinking within <think> </think> tags and a concise factual answer within <answer> </answer> tags. User: {question} Assistant:"}

\textbf{Open-Ended Answer Prompt:}

\texttt{prompt = f"You are a Medical Assistant. The User askes a question based on the provided medical image. Consider all visual and contextual details. Share your reasoning freely within <think> </think> tags, and conclude with your best judgment within <answer> </answer> tags. User: {question} Assistant:"}
\end{tcolorbox}

\section{Prompts for LLM-as-Judge}

\begin{verbatim}
evaluation_prompt = ```
Task: 

You are given a question, a ground truth, and a prediction in medical analysis. 
Evaluate the model prediction for its relevance, accuracy, and alignment with 
the ground truth. 

Reasoning Scoring: if the reasoning of the prediction is similar to ground truth, 
returning 1, otherwise returning 0.
Prediction score: if the final prediction is similar to ground truth, 
returning 1, otherwise returning 0.
Output structure:
{
    "evaluation": Provide a concise justification sentence explaining why you 
    rated the score.
    "reasoning_score": reasoning score
    "prediction_score": prediction score
}
'''
\end{verbatim}

\end{document}